%% file: main.tex
\documentclass{article}

\input{preamble}

 \usepackage[main, final]{neurips_2026}

\usepackage[utf8]{inputenc} 
\usepackage[T1]{fontenc}    
\usepackage{hyperref}       
\usepackage{url}            
\usepackage{booktabs}       
\usepackage{amsfonts}       
\usepackage{nicefrac}       
\usepackage{microtype}      
\usepackage{xcolor}         

\usepackage{xspace}
\newcommand{\modelname}{\textsc{ReToken}\xspace}

\definecolor{ava}{RGB}{150, 0, 255}

\title{ReToken: Improving Long-Context VLMs with  \\ Visual Retrieval Token}

\author{
Yao Xiao$^1$ \And
Reuben Tan$^2$ \And
Zhen Zhu$^{1,3}$\thanks{This work was done while the author was at UIUC (currently at Google DeepMind).} \And
Yuqun Wu$^1$ \AND
Jianfeng Gao$^2$ \And
Derek Hoiem$^1$   \and
$^1$University of Illinois Urbana-Champaign, $^2$Microsoft Research, $^3$Google DeepMind \\
$^1$\texttt{\{yaox11, dhoiem\}@illinois.edu}
}

\begin{document}
\maketitle

\input{text/0_abstract}

\input{tab/teaser}

\input{text/1_intro}
\input{text/2_related_work}
\input{text/3_approach}
\input{text/4_experiments}

\input{text/5_ablations}

\input{text/6_conclusion}

\medskip
{
    \small
    \bibliographystyle{plainnat-nourl}
    \bibliography{ref}
}

\input{text/7_appendix}

\newpage
\end{document}

%% file: preamble.tex
\usepackage{soul,color}
\usepackage{fontawesome}
\usepackage{tcolorbox}
\tcbuselibrary{skins,breakable}
\usepackage{booktabs}
\usepackage{enumitem}
\usepackage{arydshln}
\usepackage{xcolor}

\usepackage[numbers]{natbib}

\definecolor{citecolor}{HTML}{0071BC}
\definecolor{linkcolor}{HTML}{ED1C24}
\definecolor{urlcolor}{RGB}{180, 100, 255}

\usepackage[
    colorlinks=true, 
    citecolor=citecolor,
    linkcolor=linkcolor, 
    urlcolor=urlcolor,
]{hyperref}
\let\oldhref\href
\renewcommand{\href}[2]{\oldhref{#1}{\bfseries#2}}

\usepackage{hyperref}

\definecolor{darkorange}{HTML}{B85A24}

\sethlcolor{SeaGreen!20}
\definecolor{softgreen}{RGB}{0, 150, 0}
\definecolor{WordOrange}{HTML}{FF7043}

\usepackage{lipsum}
\usepackage{graphicx}
\usepackage{algorithm}
\usepackage{algpseudocode}
\usepackage[table,dvipsnames]{xcolor}
\usepackage{amsmath}
\usepackage{bbm}
\usepackage{subcaption}
\usepackage{orcidlink}
\usepackage[accsupp]{axessibility}
\usepackage{pifont}
\usepackage{multirow}
\usepackage{mathabx}
\usepackage{caption}
\usepackage{amssymb}
\usepackage{colortbl}  
\usepackage[utf8]{inputenc} 
\usepackage[T1]{fontenc}    
\usepackage{url}            
\usepackage{booktabs}       
\usepackage{amsfonts}       
\usepackage{nicefrac}       
\usepackage{microtype}      
\usepackage{xcolor}         
\usepackage{wrapfig}
\usepackage{listings}

%% file: text/0_abstract.tex
\begin{abstract}


Long visual contexts challenge vision-language models: performance degrades as the number of distractors grows, and processing all tokens at once can exceed GPU memory limits. We present \modelname, a single learnable embedding that extracts retrieval signals from the VLM's internal representations to select query-relevant visual tokens from the pre-filled KV cache. This enables retrieval within the answering VLM, without a separate retriever or re-encoding. Despite being trained on only a small image-QA dataset, \modelname generalizes across image and video benchmarks. On Visual Haystacks, it improves Qwen3VL-8B by 13.4 points and InternVL3.5 by 12.4 points (>20\% relative). On LVBench, it transfers zero-shot to long video and improves Qwen3VL-8B by 8.0 points. Thanks to its lightweight design, both training and long-video inference fit on a single H100. Code is available at \href{https://github.com/avaxiao/ReToken}{https://github.com/avaxiao/ReToken}.

\end{abstract}

%% file: tab/teaser.tex
\begin{figure}[H]
\centering
\begin{subfigure}[t]{0.49\textwidth}
\centering
\includegraphics[width=.9\linewidth]{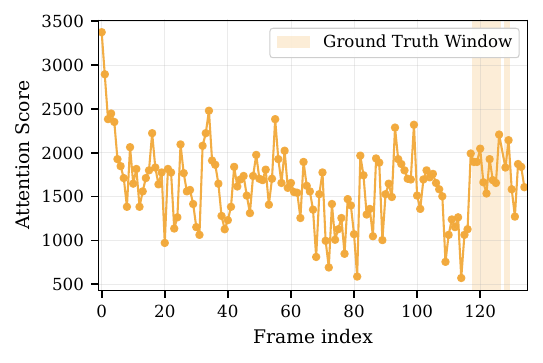}
\caption{Attention score fails to distinguish relevant frames from distractors.}
\label{fig:attn_score_one_video}
\end{subfigure}
\hfill
\begin{subfigure}[t]{0.49\textwidth}
\centering
\includegraphics[width=.9\linewidth]{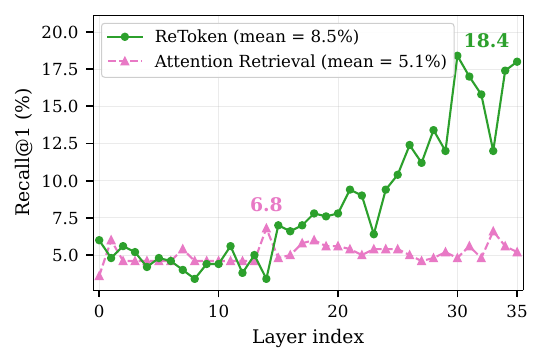}
\caption{ReToken achieves more than 3x recall in the last layer. Layer-wise VLM Recall@1 on QAEgo4D$_\text{Test-MC}$}
\label{fig:recall_per_layer_QAego4D}
\end{subfigure}
\caption{\textbf{Attention-based retrieval mechanisms are not effective for retrieving relevant visual information in VLMs}. (a) shows attention scores from question tokens (as queries) to frame tokens (as keys), across frames in one
video; (b) shows whether the top-1 retrieved frame falls within the ground-truth time window (240 candidate frames per video on average, sampled at 0.5 FPS).}
\label{fig:teaser}
\end{figure}

%% file: text/1_intro.tex
\section{Introduction}
\label{sec:intro}

Long visual contexts, e.g. from  image collections or hour-long videos, are now within the input range of vision-language models (VLMs)~\cite{liu2023visual, team2024gemma, bai2025qwen3, chen2024internvl}. However, VLMs have difficulty answering questions from long visual contexts when only a small subset of images or frames is relevant to the prompt~\cite{wu2024visual}, and sometimes processing the full context at once is computationally infeasible under GPU memory constraints. So, handling long context reduces to a retrieval problem: selecting a small subset of tokens from which the model can produce a correct answer. We ask, {\color{darkorange}\emph{does the VLM  that answers contain useful retrieval signals? If so, we can avoid adding a separate retriever model or re-encoding retrieved content for answer generation}}. When processing text, the attention signals produced by LLMs can be selective with respect to the input context, thanks to extensive long-context training~\cite{xiao2023efficient, han2024lm}. For VLMs, we find that the analogous signal is unreliable: attention between text and visual features is weakly correlated with relevance (Fig.~\ref{fig:attn_score_one_video}), and the attention-based retriever achieves only 5.1\% average recall@1 across layers on QAEgo4D$_{\text{Test-MC}}$ for Qwen3VL-8B (Fig.~\ref{fig:recall_per_layer_QAego4D}). 

Looking at what does work, we find a striking asymmetry: matching a precise target phrase against the average visual \emph{value} projections, rather than against the keys, increases recall@1 from 65.7 to 78.0 on Qwen3VL-8B and from 78.8 to 83.8 on InternVL3.5-8B in a controlled two-image setting (Tab.~\ref{tab:attention_variants}). Values carry the content that is actually propagated through attention, and they appear to provide a better space for text-based visual retrieval. However, value–value pooling does not universally outperform query–key scoring for arbitrary retrieval text. Simply averaging over all question tokens introduces noise, reversing the advantage of the value space.

Building on this finding, we propose \modelname, a single learnable embedding that is appended to the question and trained explicitly as a retrieval target. \modelname scores each frame by the cosine similarity between its projected embedding and the frame's mean value vector at the final layer, and is supervised with a class-balanced binary cross-entropy loss against ground-truth relevance labels. The token and a single projection matrix are the only added parameters; the VLM is frozen by default.

Despite its minimal footprint, \modelname yields consistent gains across
image and video benchmarks. On Visual Haystacks~\cite{wu2024visual}, it
improves Qwen3VL-8B~\cite{bai2025qwen3} by 13.4 points and InternVL3.5-8B~\cite{wang2025internvl3} by 12.4 points, corresponding to relative gains of over 20\%. More notably, although trained only on multi-image QA, \textbf{\modelname generalizes to improve performance on long-video understanding}: it transfers zero-shot to long video, yielding an 8.0-point improvement on LVBench~\cite{wang2025lvbench} with Qwen3VL-8B, where the average video length exceeds an hour. Our contributions are:
\begin{itemize}
\item We identify that retrieval scores computed in the \emph{value} space,
rather than the conventional query-key space, provide a substantially stronger signal for retrieving visual information.
\item We introduce \modelname, a lightweight learnable retrieval token that enables pretrained VLMs to identify relevant visual information more effectively.
\item We show that \modelname trained only on multi-image QA transfers
zero-shot to long-video benchmarks, suggesting a practical path toward
scalable long-context multimodal reasoning.
\end{itemize}

%% file: text/2_related_work.tex
\section{Related Work}

\textbf{Visual Retrieval}. The dominant approach to visual retrieval is to
train image-text embedding models that align visual and language feature
spaces, either with dual encoders such as CLIP~\cite{radford2021learning},
Perception Encoder~\cite{bolya2025perception}, and
SigLIP2~\cite{tschannen2025siglip}, or by adapting a VLM into a universal
embedder, as in E5-V~\cite{jiang2024e5} and LamRA~\cite{liu2025lamra}.
While effective, all of these operate as external retrievers separate from
the VLM: relevant images must first be selected by the retriever and then re-encoded by the VLM for answer generation. A related family in long-video QA chains a separate localizer with an answerer model: SeViLA~\cite{yu2023self} repurposes BLIP-2~\cite{li2023blip} as both a keyframe localizer and an answerer, while VideoAgent~\cite{wang2024videoagent} and VideoTree~\cite{wang2025videotree} build agentic pipelines that iteratively retrieve and caption keyframes for an LLM. A second line of work instead treats the model's own attention scores as a retrieval signal, unifying retrieval and generation within a single forward pass. InfLLM~\cite{xiao2024infllm} and EM-LLM~\cite{fountas2025human} demonstrate this effectively in NLP by selecting key-value blocks based on query-to-key attention, and ReKV~\cite{di2025streaming} transfers the mechanism to the visual domain without modification. We diagnose the limitations of attention-based retrieval in VLMs and propose \modelname, which performs retrieval in the value space via a learnable token.

\textbf{Long Context Understanding.} When handling long visual contexts, prior work follows three main directions. Memory-based methods such as MovieChat~\cite{song2024moviechat} and MA-LMM~\cite{he2024ma} maintain a fixed-size memory bank over streaming input, merging or evicting older content to bound state. Token compression methods such as Chat-UniVi~\cite{jin2024chat}, LLaMA-VID~\cite{li2024llama}, LongVU~\cite{shen2024longvu}, and Video-XL~\cite{shu2025video} prune, merge, or summarize visual tokens before the LLM to shorten the prefix and reduce attention and KV-cache cost. Both families bound the visual context independently of the query. Retrieval-based methods instead defer context selection to inference time, picking a query-relevant subset to preserve fine-grained evidence: Goldfish~\cite{ataallah2024goldfish} chunks the video into clips and retrieves the top-$K$ by caption-query similarity; Video-RAG~\cite{luo2024video} augments retrieval with visually-aligned auxiliary text from ASR, OCR, and object detectors; and DrVideo~\cite{ma2025drvideo} converts the video into a long
document and retrieves question-relevant passages for the LLM. End-task accuracy in this family, however, is bottlenecked by the retriever rather than the
LLM, and our work targets this bottleneck directly.

\textbf{Learnable Tokens for Visual Aggregation.} \modelname is most directly related to methods that introduce learnable tokens into a VLM to aggregate or retrieve visual content. The Q-Former in BLIP-2~\cite{li2023blip} and the Perceiver Resampler in Flamingo~\cite{alayrac2022flamingo} use a small set of learnable queries to compress variable-length visual features into a fixed-size representation for a frozen LLM, and InstructBLIP~\cite{dai2023instructblip} extends Q-Former to be text-conditioned. These modules are trained jointly with vision-language alignment as part of the bridge between encoder and LLM, and produce many tokens (typically 32 or 64) intended to carry the visual content forward into generation. SPRING~\cite{zhu2025one} prepends pluggable soft-prompt tokens to externally retrieved passages to help a frozen text-only LLM consume them. The visual summarization token in Video-XL~\cite{shu2025video} similarly compresses a chunk of visual KVs into a single token. \modelname differs along three axes: (i) \textit{operational mechanism}. Q-Formers and prompt tuning use learnable tokens as static, query-agnostic input conditions. In contrast, the ReToken is generated dynamically. A first-pass placeholder produces an output token that retrieves visual KV contexts relevant to the query. (ii) \textit{supervision}. Q-Formers are trained as visual compressors under alignment or generation losses, while \modelname is supervised by an explicit retrieval loss scored directly against final-layer value vectors. (iii)
\textit{capacity}. Our method needs only a single token rather than a set of 32--64. The second half of the contribution is diagnostic: attention (query–key) scores are unreliable for visual retrieval in pretrained VLMs, while the value space carries a much stronger signal.

%% file: text/3_approach.tex
\section{Approach}

In this section, we begin by analyzing the limitations of attention-based retrieval methods for identifying relevant visual tokens in Sec.~\ref{subsec:attention_based_retrieval}. Motivated by these insights, we then  discuss our proposed \modelname approach in Sec.~\ref{subsec:retoken}, which uses a learnable token embedding to help compute more informative retrieval scores over relevant visual tokens.

\subsection{Attention-Based Retrieval}  
\label{subsec:attention_based_retrieval}  

In our empirical setting, we consider a VLM based on a decoder-only LM comprising $N$ transformer layers $\{L_1, \dots, L_N\}$. Given input question tokens $\mathbf{T}_t$ and visual tokens $\mathbf{I}_v$, the VLM autoregressively generates the response $\mathbf{y}$ based on the conditional probability $p(\mathbf{y} \mid \mathbf{I}_v, \mathbf{T}_t)$. For multi-image and video inputs, the visual tokens are partitioned into $F$ frames, $\mathbf{I}_v = \{\mathbf{I}_v^{(f)}\}_{f=1}^{F}$, with each frame contributing $M/F$ tokens (where $M$ is the total number of visual tokens). For long videos, $M$ can be prohibitively large, rendering full-context inference infeasible. Thus, our goal is to retrieve a subset of the $K$ most relevant frames, where $K \ll F$, whose tokens suffice to answer the question.

\textbf{Attention Scores as a Retrieval Signal}. At each layer $l \in \{1, 2, \dots, N\}$, the VLM computes attention scores $\mathbf{A}^{(l)} = \mathbf{Q}^{(l)} \mathbf{K}^{(l)\top} / \sqrt{d}$, which indicate how much each token contributes to the final response. Tokens with higher attention contributions can be interpreted as being more relevant to the question. For our analysis, we use the state-of-the-art ReKV~\cite{di2025streaming} approach that directly uses attention scores as the retrieval scores and aggregates them at the frame level. At the $l-$th layer, we compute the mean key vector of each frame's visual tokens and the mean query vector over the question tokens as:
\begin{equation}
\bar{\mathbf{k}}^{(l)}_f = \frac{1}{|\mathbf{I}_v^{(f)}|} \sum_{i \in \mathbf{I}_v^{(f)}} \mathbf{k}^{(l)}_i,
\qquad
\bar{\mathbf{q}}^{(l)} = \frac{1}{E} \sum_{t=1}^{E} \mathbf{q}^{(l)}_t
\end{equation}
where $E$ denotes the total number of question tokens. Then, the frame-level retrieval score $\mathbf{s}_f^{(l)}$ and top-$K$ frame selection are computed as: 
\begin{equation}
\mathbf{s}_f^{(l)} = \bar{\mathbf{q}}^{(l)\top} \bar{\mathbf{k}}^{(l)}_f,
\qquad
\mathcal{S}_K^{(l)} = \mathrm{TopK} (\ \mathbf{s}_f^{(l)} )
\label{eq:attn-retrieval}
\end{equation}
where $\mathcal{S}_K^{(l)}$ denotes the $K$ frames most relevant to the question at layer $l$. Given the per-layer retrieval sets $\{\mathcal{S}_K^{(l)}\}_{l=1}^{N}$, the model re-runs generation through the frozen VLM with \emph{layer-wise} sparse attention: at each layer $l$, the question and answer tokens attend only to the visual tokens belonging to $\mathcal{S}_K^{(l)}$. The reduced visual context at layer $l$ is therefore $[\mathbf{I}_v[\mathcal{S}_K^{(l)}]]$, and the answer is decoded autoregressively over this layer-dependent context $[\mathbf{I}_v[\mathcal{S}_K^{(l)}], \mathbf{T}_t]$.

\begin{figure*}[t!]
    \centering
    \hspace*{-2em}
    \makebox[\textwidth][c]{%
        \includegraphics[width=1\textwidth]{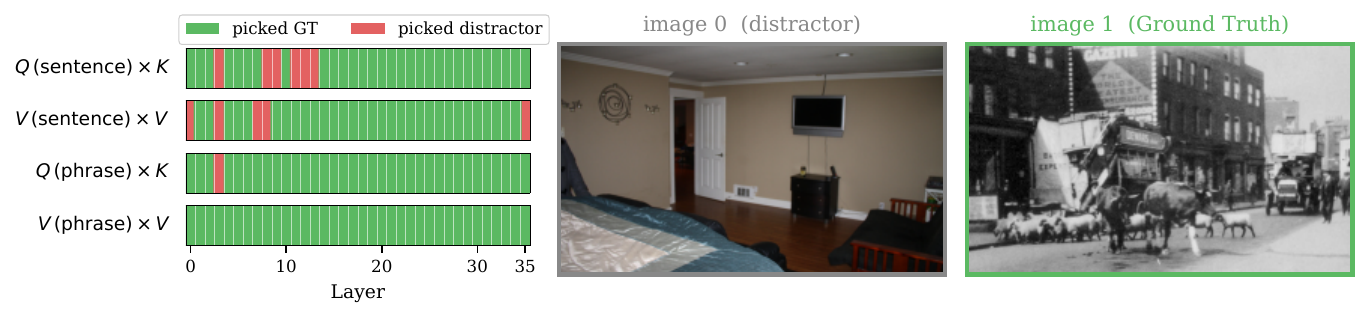}
    }
    \caption{\textbf{Retrieval Example}. Question/Sentence: "For the image with a cow, is there a truck?" Target phrase: "a cow".}
    \label{fig:vhs_example}
\end{figure*}

\textbf{Limitations of Attention-Based Retrieval}. While attention-based retrieval is intuitive, two structural issues motivate our approach. First, attention is trained for next-token prediction rather than retrieval, so high-attention tokens are not guaranteed to correspond to query-relevant content. This mismatch is exacerbated by the composition of typical VLM training data: the input images or videos are almost always fully relevant to the question, so the model is never required to \emph{select} among visual inputs. The consequences are visible in both image and video settings: on Visual Haystacks~\cite{wu2024visual} the layer-wise $\bar{\mathbf{q}}^\top \bar{\mathbf{k}}_f$ retriever lands at $63.3\%$ recall@1 even in the simplest two-image case (Tab.~\ref{tab:attention_variants}), and on long-video QAEgo4D$_{\text{Test-MC}}$ it averages only $5.1\%$ recall@1 across layers (Fig.~\ref{fig:recall_per_layer_QAego4D}). Second, averaging the question tokens to form a single query is itself a heuristic, and the resulting rankings shift with the choice of retrieval text. As shown in Tab.~\ref{tab:attention_variants}, replacing the full question sentence with a target phrase that directly names the entity of interest improves Recall@1. Fig.~\ref{fig:vhs_example} illustrates a question sentence and its target phrase.

\begin{wraptable}[12]{r}{0.5\textwidth}
    \vspace{-1.2em}
    \caption{\textbf{$\text{Value} \times \text{Value}$ is more informative, but it is sensitive to the input}. Recall@1 with 2 input images, retrieving 1 image based on the retrieval score.}
    \label{tab:attention_variants}
    \centering
    \setlength\tabcolsep{2pt}
    \resizebox{\linewidth}{!}{
        \begin{tabular}{@{}l | cc | cc@{}}
        \toprule
        Retrieval Text  &  $\bar{\mathbf{q}}^{\top} \bar{\mathbf{k}}_f$ &  $\bar{\mathbf{v}}^{\top} \bar{\mathbf{v}}_f$ &  Qwen3VL &  InternVL3.5  \\
        \cmidrule{1-5}
         \multirow{2}{*}{Sentence}  &  \checkmark  &  & 63.3 & 78.5  \\
           &   & \checkmark  & 62.6 & 75.6 \\
        \noalign{\vskip 3pt}\cdashline{1-5}\noalign{\vskip 3pt}
         \multirow{2}{*}{Target Phrase}  &  \checkmark  &  &  65.7 & 78.8 \\
           &    & \checkmark  & \textbf{78.0} & \textbf{83.8} \\
        \bottomrule
        \end{tabular}
    }
\end{wraptable}

\textbf{Why Values Carry Retrieval Signal.} Within a transformer attention layer, the value of a token carries the content propagated to any token that attends to it, while the query-key inner product only determines how that content is aggregated. Pooling the value projections within each frame therefore yields a representation of the content the frame contributes to attending tokens, making value features more sensitive to the retrieval text than key features. Tab.~\ref{tab:attention_variants} confirms this consistently across both backbones: with a precise target phrase that names the entity of interest, value-space pooling identifies the ground-truth image at $78.0\%$ recall@1 vs.\ $65.7\%$ for the corresponding query-key score on Qwen3VL-8B, and at $83.8\%$ vs.\ $78.8\%$
on InternVL3.5-8B. This mirrors observations in visual segmentation, where value features are reported to be more informative while query-key features can be replaced by alternative aggregation signals~\cite{zhou2022extract, wang2024sclip, lan2024proxyclip}; TextRegion~\cite{xiao2025textregion} in particular shows that, in image-text models, value features in the final attention block are rich in visual-language semantics, whereas attention weights primarily serve
as an aggregation mechanism. This sensitivity cuts both ways, however: with the full question sentence, averaging mixes in many uninformative tokens, and the resulting noisy query erases or even reverses the value-space advantage. The value space thus rewards a precise retrieval target and penalizes a noisy one.

\subsection{ReToken: One Token for Visual Retrieval}
\label{subsec:retoken}

To improve retrieval, we introduce \modelname, a single learnable embedding $\mathbf{X}_r \in \mathbb{R}^{d}$ that is appended to the question and trained explicitly as a retrieval target. \modelname addresses the rephrasing instability by replacing the question-averaged query with a token \emph{learned} from data, and strengthens the retrieval score by computing in the value space rather than the query--key space.

\textbf{Retrieval Score}. Given an input sequence with visual tokens $\mathbf{I}_v$, question tokens $\mathbf{T}_t$, and the appended retrieval token $\mathbf{X}_r$, we compute retrieval scores at the final layer $L_N$ via a lightweight projection $\mathbf{W}_r \in \mathbb{R}^{d \times d}$, which together with $\mathbf{X}_r$ constitutes the only added parameters. The retrieval score for the $f$-th frame is the cosine similarity between the projected embedding and the frame's mean value vector $\bar{\mathbf{v}}_f^{(N)}$,
\begin{equation}
\bar{\mathbf{v}}_f^{(N)} = \frac{1}{|\mathbf{I}_v^{(f)}|} \sum_{i \in \mathbf{I}_v^{(f)}} \mathbf{v}_i^{(N)},
\quad
\mathbf{Z}_r=\mathbf{W}_r\,\mathbf{X}_r^{(N)},
\quad
s_f^{(N)} = \cos\!\left( \mathbf{Z}_r,\; \bar{\mathbf{v}}_f^{(N)} \right)
\label{eq:retoken-score}
\end{equation}
For inference, we compute $\mathcal{S}_K^{(N)} = \mathrm{TopK}(s_f^{(N)})$ once at the final layer $L_N$ and broadcast it to all layers, rather than maintaining per-layer subsets $\{\mathcal{S}_K^{(l)}\}$. We do not apply this at training time, since our training data is short-context and retrieval is unnecessary.

\begin{figure*}[t!]
    \centering
    \makebox[\textwidth][c]{%
        \includegraphics[width=\textwidth]{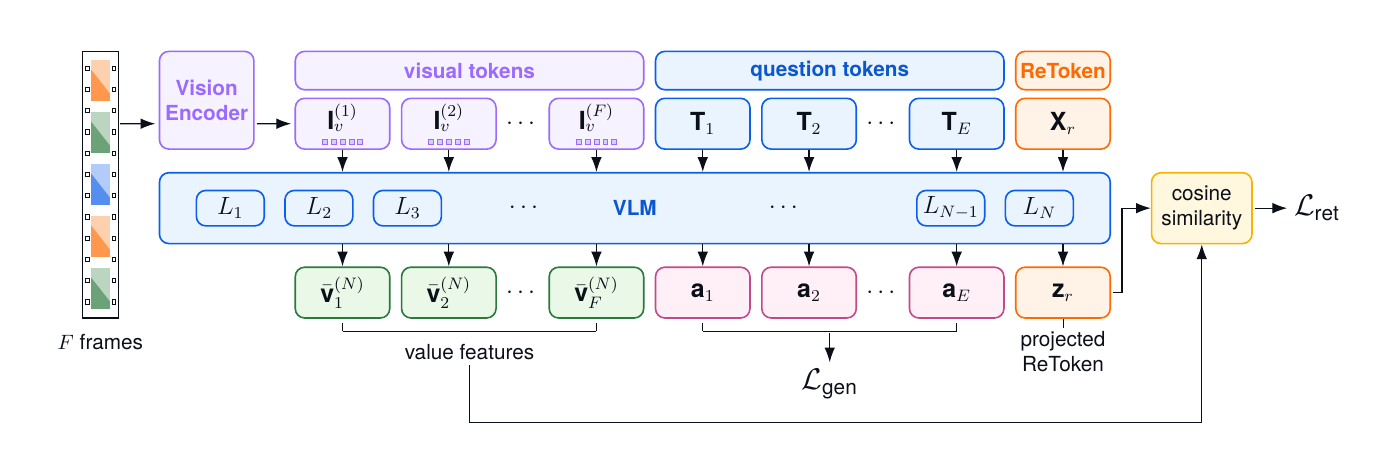}
    }
    \caption{\textbf{Training Pipeline.} The visual tokens, question tokens, and a learnable embedding (\modelname) are fed into the LLM decoder. The \modelname output from the final layer serves as a retrieval query, scoring each frame based on its cosine similarity to the frame’s pooled value features. The resulting scores are supervised using ground-truth frame-relevance labels and a class-balanced binary cross-entropy loss. By default, we train \modelname while keeping the VLM frozen. The generation loss is used only in the partial VLM fine-tuning setting.}
    \label{fig:training}
\end{figure*}

\textbf{Training}. We train \modelname with the VLM kept frozen by default; only the retrieval token $\mathbf{X}_r$ and a single final-layer projection $\mathbf{W}_r$ are updated. We supervise these parameters with a \emph{retrieval loss} $\mathcal{L}_{\text{ret}}$ that compares the final-layer retrieval scores against ground-truth relevance labels $y_f \in \{0,1\}$. 

Let $\mathcal{F}^{+} = \{f : y_f = 1\}$ and $\mathcal{F}^{-} = \{f : y_f = 0\}$ denote the sets of relevant and irrelevant images, respectively. To prevent the loss from being dominated by the irrelevant images, we apply a class-balanced binary cross-entropy in which the positive and negative terms are averaged separately,
\begin{equation}
\mathcal{L}_{\text{ret}} = - \frac{1}{|\mathcal{F}^{+}|} \sum_{f \in \mathcal{F}^{+}} \log \sigma(\tau s_f^{(N)}) \;-\; \frac{1}{|\mathcal{F}^{-}|} \sum_{f \in \mathcal{F}^{-}} \log \sigma(- \tau s_f^{(N)})
\label{eq:retoken-bce}
\end{equation}
where $s_f^{(N)}$ is the retrieval score for image $f$ at the final layer $L_N$, $\sigma(\cdot)$ is the sigmoid, and $\tau$ is a learnable logit scale parameter.

We additionally explore a \emph{partial-tuning} variant in which the early layers of the VLM are tuned to give the visual representation more flexibility. In this setting, training is driven by two complementary losses, illustrated in Fig.~\ref{fig:training}. In addition to the retrieval loss above, we use a \emph{generation loss} $\mathcal{L}_{\text{gen}}$, the standard next-token prediction loss on the answer, which preserves the model's question-answering ability and prevents the unfrozen layers from drifting under $\mathcal{L}_{\text{ret}}$ alone. The total loss is
\begin{equation}
\mathcal{L} = \mathcal{L}_{\text{ret}} + \lambda \, \mathcal{L}_{\text{gen}}
\end{equation}

\begin{figure*}[t!]
    \centering
    \makebox[\textwidth][c]{%
        \includegraphics[width=\textwidth]{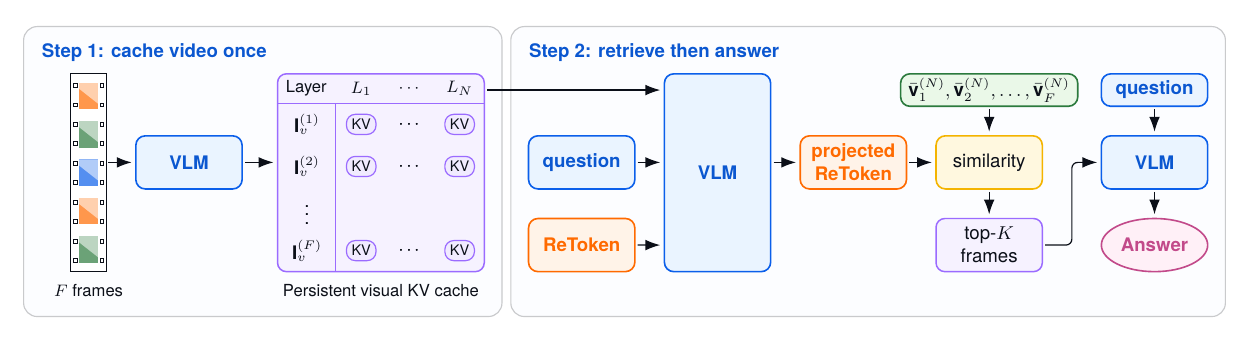}
    }
    \caption{\textbf{Inference Pipeline}. Our method first caches features for all input frames. Given a text query, ReToken retrieves the relevant frames, and the answer is generated from their cached features.}
    \label{fig:inference}
\end{figure*}

\textbf{Inference}. At test time, we use a two-pass \emph{retrieve-then-answer} pipeline (Fig.~\ref{fig:inference}). The video is first encoded once, and each question then retrieves a subset of the visual KV cache for answer generation. 

Cache video once: At video ingestion time, the VLM processes the visual tokens and stores the resulting per-layer KV cache in a persistent cache. For short videos, this is a single full-context forward pass. For long videos, we follow ReKV~\cite{di2025streaming} and encode the video chunk by chunk with a sliding-window attention mask: each chunk attends only to the most recent $l_m$ visual tokens, while its newly produced KV states are appended to the cache. Older KV states can be offloaded to CPU memory when GPU memory is limited.

Retrieve then answer: Given a question, we run two forward passes through the frozen VLM over the cached visual context. The first pass computes the retrieval set $\mathcal{S}_K^{(N)}$ from the final layer; the second pass generates the answer conditioned on the selected frames. We use two passes because \modelname is supervised only at $L_N$ (Eq.~\ref{eq:retoken-bce}). The first pass enables the retrieval token to fully integrate textual and visual content, while the second pass enables attention to a consistent set of visual tokens, matching how it was trained.

\emph{First pass -- retrieval.}
We append the retrieval token $\mathbf{X}_r$ to the question and run the first pass over the cached visual KV. For short videos $\mathbf{X}_r$
attends to all cached visual tokens at layer $l < N$, and the index
$\mathcal{S}_K^{(N)}$ is computed at $L_N$ via Eq.~\ref{eq:retoken-score}.
For long videos the early-layer attention budget cannot accommodate the
full cache, so at each layer $l < N$ we restrict $\mathbf{X}_r^{(l)}$'s
attention as follows: (1) project the contextualized $\mathbf{X}_r^{(l)}$
through that layer's frozen value projection; (2) score every frame by
the cosine similarity between this projection and the frame's mean value
vector $\bar{\mathbf{v}}_f^{(l)}$ at layer $l$; (3) restrict
$\mathbf{X}_r^{(l)}$'s attention to the top-$K'$ frames under this score
before computing layer $l$'s output. We use $K' = 256$ by default, which
keeps each early-layer attention well within GPU memory while leaving
headroom for the final-layer ranking to refine the selection. At $L_N$
we then compute Eq.~\ref{eq:retoken-score} over the per-frame value means
to obtain $\mathcal{S}_K^{(N)}$ , reordered by original
timestamp.

We note an asymmetry between this retrieval-pass budget $K'$ and the answer-stage budget $K$: \modelname's retrieval pass attends to up to $K' = 256$ frames per early layer, which is different from the $K$ used at answer time. \modelname therefore has access to more visual context when \emph{ranking} candidates, while it still uses only $K$-frame visual context when \emph{generating} the answer.

\emph{Second pass -- answer.}
For each layer $l$, we load only the visual KV cache belonging to frames in $\mathcal{S}_K^{(N)}$, and run a standard
generation pass over this reduced visual context together with the
question. The answer stage therefore attends to about $K \cdot M / F$
visual tokens instead of all $M$, while the expensive video encoding is
performed only once per video. This is particularly efficient when
multiple questions are asked about the same video. Full prompt templates
and a worked example of how \modelname interacts with the contextual
visual tokens and the question are provided in
Supp.~\ref{supp:supp_prompts}.

%% file: text/4_experiments.tex
\section{Experiments}


\subsection{Settings}
\label{sec:implementation_details}

\textbf{Implementation Details}. We use the greedy decoding configuration for \modelname and every baseline. We train \modelname using \texttt{Qwen3VL-8B} and \texttt{InternVL3.5-8B} on an image question-answering (QA) dataset. For the default frozen-VLM setting, we use an effective batch size of 64 and a learning rate of $3 \times 10^{-4}$ to train the learnable token for 3 epochs on a single H100 GPU with \texttt{Qwen3VL-8B}. Since \texttt{InternVL3.5-8B} requires more tokens to represent an image, resulting in substantially higher computational and memory costs, we train \texttt{InternVL3.5-8B} for only one epoch. We adopt a linear warmup schedule followed by cosine learning rate decay. For the partial fine-tuning setting, we use a learning rate of $2 \times 10^{-5}$ and an effective batch size of 64. As this setting converges rapidly and begins to overfit after one epoch, we apply early stopping at the end of the first epoch. Training \texttt{Qwen3VL-8B} under this setting takes approximately 4 hours on a single H100 GPU.

For long-video inference, we use an encoding chunk size of 128 frames and a sliding-window length of $l_m{=}30{,}000$ tokens, which corresponds to roughly 153 frames at 196 tokens per frame. 

\textbf{Training Datasets}. Our default multi-image training dataset comes from the MIRAGE~\cite{wu2024visual} fine-tuning dataset, whose examples are each annotated with a relevant/irrelevant label per image. We use a 95\%/5\% train/validation split. It is a multi-image QA (MIQA) dataset that combines existing MIQA datasets (RetVQA~\cite{penamakuri2023answer}, SlideVQA~\cite{tanaka2023slidevqa}, and WebQA~\cite{chang2022webqa}) with synthetic MIQA data adapted from the LLaVA Visual Instruct 150K dataset~\cite{liu2024improved} via clustering and distractor sampling. Please refer to Supp.~\ref{supp:training_data} for more details.

\textbf{Evaluation Datasets}. 
We evaluate \modelname on four benchmarks chosen to probe complementary aspects of retrieval and long-context understanding. 

\emph{Visual Haystacks (VHs)}~\cite{wu2024visual} is a benchmark introduced alongside MIRAGE to evaluate the pure visual recognition ability of vision-language models. Constructed from the COCO dataset~\cite{lin2014microsoft}, it consists of 1{,}000 question-answer pairs and provides query-relevant image annotations. The answer is always either ``Yes'' or ``No''. For each QA example, a varying number of distractor images is added to test the model's ability to locate informative content within a large pool of inputs. 

\emph{QAEgo4D$_\text{Test-MC}$}~\cite{di2024grounded} is the multiple-choice subset of the QAEgo4D-test benchmark~\cite{9857465}, focusing on question answering over long egocentric videos. Each example is annotated with the video segments relevant to the question. Video lengths range from 4 to 20 minutes.

\emph{LVBench}~\cite{wang2025lvbench} is a benchmark for extreme long-video understanding, comprising 103 publicly sourced YouTube videos totaling roughly 117 hours, with an average length of 68 minutes and individual videos extending up to 2 hours. This is a multiple-choice dataset, and the metric is accuracy.

\emph{Video-MME}~\cite{fu2025video} consists of 900 videos and 2{,}700 question-answer pairs. Video durations range from 11 seconds to 1 hour, partitioned into short ($<$2 min), medium (4--15 min), and long (30--60 min) splits. Each question is multiple-choice, and we report accuracy as the evaluation metric.

\subsection{Visual Haystacks Retrieval and Accuracy}

\input{tab/visual_haystack_retrieval}

We compare performance on the single-needle task of Visual Haystacks~\cite{wu2024visual}, where exactly one image in the haystack is relevant to the query. The context size $C$ is the number of images in the haystack, with larger $C$ introducing more distractors, while $K$ is the number of retrieved images.

\textbf{Retrieval Behavior of ReToken and Attention-Based Retrieval}. Fig.~\ref{fig:accuracy_vs_topN} compares \modelname and the attention-based retriever, ReKV~\cite{di2025streaming}, across retrieval budgets from $K{=}1$ to $K{=}128$ at $C{=}50$ and $C{=}100$. When $K >= C$, no retrieval is needed since all images are used as input. The two methods exhibit opposite trends in the meaningful retrieval regime ($K < C$): \modelname performs best at small $K$ and degrades as more images (mostly distractors) are retrieved, whereas ReKV starts low and improves with larger $K$. The crossover reflects a precision--recall tradeoff: \modelname is precise per slot, while ReKV needs a wider budget to recall the relevant image. 

\textbf{Retrieval Strategy Comparison}. We also compute recall to measure the retrieval ability of different strategies on Visual Haystacks. Here, recall measures whether the retriever successfully selects the ground-truth image when $K{=}1$. Tab.~\ref{tab:visual_haystack_retrieval} compares \modelname against several retrieval strategies on Visual Haystacks. We compare against four baselines: \emph{Standard} pre-fills all images (the default VLM inference pipeline); \emph{GT Cache} uses the ground-truth image's KV cache as input, serving as an oracle upper bound; \emph{SigLIP2}~\cite{tschannen2025siglip} first retrieves a single image with \texttt{siglip2-giant-opt-patch16-384} and feeds its KV cache to the VLM; and \emph{CoT} prompts the VLM to first generate a target search phrase, then uses that phrase as the query for ReKV attention-based retrieval (full prompt in Supp.~\ref{supp:cot_prompts}).

\modelname substantially outperforms all baselines, retaining most of the GT Cache upper-bound accuracy at large context size ($C=50$) while the others fail. Observing the accuracy for GT Cache in Tab.~\ref{tab:visual_haystack_retrieval}, we notice an interesting phenomenon: the accuracy degrades with higher C, even though only the cache for the same single ground truth relevant image is used in each case (86.5 for $C=2$ vs. 80.7 for $C=50$). This is because the image tokens have attended to previous images, which are irrelevant in this case, picking up distracting information. While harmful in this particular setting, the attention across frames is likely helpful in many video contexts. Refer to Supp.~\ref{supp:KV_cache_distraction}.

\input{tab/visual_haystack}

\textbf{Accuracy Comparison}. Tab.~\ref{tab:visual_haystack} reports accuracy across context sizes from $C{=}1$ (only the query-relevant image, no distractors) to $C{=}100$. The retrieval budget $K$ is set to 1. At $C{=}1$, \modelname matches the vanilla \texttt{Qwen3VL-8B} backbone, as we skip retrieval when $C<=K$. As the context size grows, the gap between \modelname and the vanilla backbone widens substantially. The one exception is InternVL3.5-8B at $C{=}100$, which we attribute to its 1-epoch training budget (Sec.~\ref{sec:implementation_details}). \modelname also compares favorably against external baselines, surpassing all baselines across all context sizes, including the dedicated multi-image RAG framework MIRAGE. 


\subsection{Image-to-Video Transfer Evaluation}

\begin{wraptable}[7]{r}{0.42\textwidth}
\centering
\vspace{-2.5em}
\caption{Zero-shot performance on QAEgo4D$_\text{Test-MC}$.}
\label{tab:accuracy_QAEgo4D}
\setlength\tabcolsep{3pt}
\begin{tabular}{@{}c | ccc@{}}
\toprule
$K$  & Uniformly & ReKV & \textbf{\textsc{ReToken}} \\

\midrule
1  & 42.8 & 42.6 & \textbf{49.6} \\
16 & 53.6 & 57.8 & \textbf{60.0} \\
32 & 55.2 & 59.8 & \textbf{60.6} \\


\bottomrule
\end{tabular}
\end{wraptable}

Having validated \modelname on images, we next evaluate its transfer to video. Notably, all results in this section use \modelname trained \emph{only} on the multi-image MIRAGE training data, demonstrating strong zero-shot transfer. We evaluate the frozen \texttt{Qwen3VL-8B} on videos sampled at 0.5 FPS. ``Uniformly'' denotes loading the KV cache from $K$ uniformly sampled frames as input.

\textbf{Accuracy on QAEgo4D$_\text{Test-MC}$}. Tab.~\ref{tab:accuracy_QAEgo4D} shows that \modelname's advantage is most pronounced under tight retrieval budgets. With $K{=}1$, \modelname achieves a 6.8-point improvement over uniform sampling, while ReKV actually performs slightly worse than uniform sampling. 
Unlike multi-image QA where images are independent, neighboring video frames provide complementary context, so accuracy continues to climb as $K$ grows even where \modelname's lead over uniform sampling narrows.

\textbf{Long Video Understanding}. The benefits of retrieval grow with video length (Tab.~\ref{fig:videomme_gain}). On the Short split of Video-MME, \modelname provides no gain, since these videos last less than 2 minutes (60 frames at 0.5 FPS), which falls below answer-stage frame budget of $K=100$ and therefore does not trigger retrieval. As video length grows, however, the gain becomes substantial. \modelname achieves an 8.0-point improvement on LVBench (Tab.~\ref{tab:long_video_benchmarks}).

\input{tab/sota_video_benchmarks}



%% file: tab/visual_haystack_retrieval.tex
\begin{figure*}[b!]
\centering
\begin{minipage}[c]{0.42\textwidth}
    \vspace{0pt}
    \hspace*{-2em}
    \centering
    \makebox[\textwidth][c]{%
        \includegraphics[height=.185\textheight]{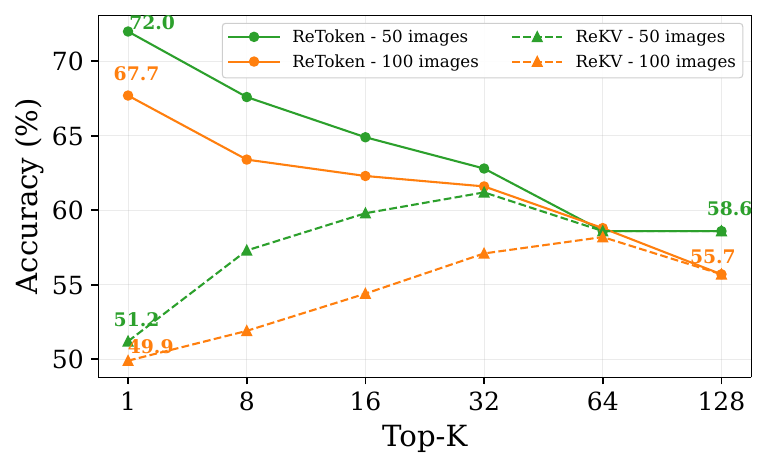}
    }
    \captionof{figure}{\textbf{Accuracy vs. Retrieval  \\ Budget $K$}. Frozen \texttt{Qwen3VL-8B}.}
    \label{fig:accuracy_vs_topN}
\end{minipage}
\hfill
\begin{minipage}[c]{0.57\textwidth}
    \captionof{table}{\textbf{Retrieval Strategy Comparison.} We compare \texttt{ReToken} against different retrievers on Visual Haystacks with retrieval budget $K{=}1$. {\sethlcolor{lightgray!40}\hl{“GT Cache”}} uses the KV cache of the ground truth image as input and serves as the oracle upper bound for any retriever. All results are based on \texttt{Qwen3VL-8B} with the VLM frozen.}
    \label{tab:visual_haystack_retrieval}
    \centering
    \setlength\tabcolsep{1.5pt}
    \resizebox{\linewidth}{!}{%
    \begin{tabular}{@{}lcc|ccc|c@{}}
    \toprule
    Metric
    & Standard
    & GT Cache
    & SigLIP2
    & ReKV
    & CoT 
    & \textbf{\textsc{ReToken}} \\
    \midrule
    \multicolumn{3}{@{}l}{\textbf{\textcolor{gray}{$C=2, K=1$}}} \\
    Recall   & N/A  & 100  & 76.4 & 63.3 & 65.3 &  \textbf{88.5} \\
    Accuracy & 82.0 & 86.5 & 82.8 & 73.0 & 77.8 & \textbf{85.9} \\
    \midrule
    \multicolumn{3}{@{}l}{\textbf{\textcolor{gray}{$C=50, K=1$}}} \\
    Recall   & N/A  & 100  & 20.8 & 1.8  & 3.1  & \textbf{64.7} \\
    Accuracy & 58.6 & 80.7 & 60.7 & 51.2 & 51.5 & \textbf{72.0} \\
    \bottomrule
    \end{tabular}
    }
\end{minipage}
\end{figure*}













%% file: tab/visual_haystack.tex
\begin{table*}[t!]
\caption{\textbf{\texttt{ReToken}'s advantage grows with context size, yielding over 20\% relative gain at C=50, when freezing the VLM.} Performance on Visual Haystacks. $C$ denotes the number of context images (i.e., images provided as input). $E$ indicates a context overflow or CUDA out-of-memory error.}
\label{tab:visual_haystack}
\setlength\tabcolsep{2pt}
\centering
\resizebox{1\textwidth}{!}{
\begin{tabular}{@{}lcccccccc@{}}
\toprule
Method  & $C=1$ & $C=2$ & $C=3$ & $C=5$ & $C=10$ & $C=20$ & $C=50$ & $C=100$   \\

\midrule

\multicolumn{9}{@{}l}{\textbf{\textcolor{gray}{\texttt{Generalist}}}} \\

Gemini-1.5 Pro~\cite{team2024gemini}  & 88.4 & 82.0 & 78.3 & 76.0 & 71.9 & 68.6 & 62.8 & 57.4  \\

InternVL2~\cite{chen2024expanding} & 88.1 & 80.5 & 72.3 & 63.9 & 58.8 & 55.2 & $E$ & $E$ \\

\midrule

\multicolumn{9}{@{}l}{\textbf{\textcolor{gray}{\texttt{Specialist}}}} \\

SigLIP2~\cite{tschannen2025siglip} & 72.0 & 69.2 & 68.1 & 65.3 & 64.1 & 60.3 & 58.7 & 58.3 \\


MIRAGE~\cite{wu2024visual} & 83.2 & 77.8 & 76.6 & 72.8 & 70.5 & 66.0 & 63.6 & 62.0  \\

REN$_\text{DINO·SigLIP2}$~\cite{khosla2025ren} & 81.2 & 78.6 & 77.4 & 76.0 & 74.0 & 72.1 & 68.3 & 65.5 \\

\midrule
\multicolumn{9}{@{}l}{\textbf{\textcolor{gray}{\texttt{Freezing the VLM}}}} \\

Qwen3VL-8B~\cite{bai2025qwen3}  & \textbf{87.8} & 82.0 & 77.9 & 74.7 & 69.2 & 64.0 & 58.6 & 55.7  \\

\rowcolor{SeaGreen!20}
\quad \textbf{+ \textsc{ReToken}} &  \textbf{87.8 {\color{softgreen}\text{\small(+0.0)}}} & \textbf{85.9 {\color{softgreen}\text{\small(+3.9)}}} & \textbf{84.4 {\color{softgreen}\text{\small(+6.5)}}} & \textbf{82.5 {\color{softgreen}\text{\small(+7.8)}}} & \textbf{80.7 {\color{softgreen}\text{\small(+11.5)}}} & \textbf{76.8 {\color{softgreen}\text{\small(+12.8)}}} & \textbf{72.0 {\color{softgreen}\text{\small(+13.4)}}} & \textbf{67.7 {\color{softgreen}\text{\small(+12.0)}}} \\

InternVL3.5-8B~\cite{wang2025internvl3} & \textbf{87.5} & 81.6 & 79.3 & 73.1 & 69.0 & 65.0 & 57.3 & 55.3 \\
\rowcolor{SeaGreen!20}
\quad \textbf{+ \textsc{ReToken}} & \textbf{87.5 {\color{softgreen}\text{\small(+0.0)}}} & \textbf{84.1 {\color{softgreen}\text{\small(+2.5)}}} & \textbf{83.7 {\color{softgreen}\text{\small(+4.4)}}} & \textbf{80.7 {\color{softgreen}\text{\small(+7.6)}}} & \textbf{76.9 {\color{softgreen}\text{\small(+7.9)}}} & \textbf{73.0 {\color{softgreen}\text{\small(+8.0)}}} & \textbf{69.7 {\color{softgreen}\text{\small(+12.4)}}} & \textbf{59.7 {\color{softgreen}\text{\small(+4.4)}}} \\

\noalign{\vskip 3pt}\cdashline{1-9}\noalign{\vskip 3pt}

\multicolumn{9}{@{}l}{\textbf{\textcolor{gray}{\texttt{Tuning the first 3 layers of the VLM}}}} \\

Qwen3VL-8B~\cite{bai2025qwen3}  & \textbf{88.5} & 85.6 & 82.4 & 79.6 & 76.5 & 71.2 & 64.2 & 61.5  \\

\rowcolor{SeaGreen!20}
\quad \textbf{+ \textsc{ReToken}} &  \textbf{88.5 {\color{softgreen}\text{\small(+0.0)}}} & \textbf{86.5 {\color{softgreen}\text{\small(+0.9)}}} & \textbf{85.2 {\color{softgreen}\text{\small(+2.8)}}} & \textbf{84.3 {\color{softgreen}\text{\small(+4.7)}}} & \textbf{82.8 {\color{softgreen}\text{\small(+6.3)}}} & \textbf{79.8 {\color{softgreen}\text{\small(+8.6)}}} & \textbf{75.0 {\color{softgreen}\text{\small(+10.8)}}} & \textbf{70.6 {\color{softgreen}\text{\small(+9.1)}}} \\

\bottomrule 
\end{tabular}
}
\end{table*}

%% file: tab/sota_video_benchmarks.tex
\begin{table*}[t!]
\caption{\textbf{ReToken helps more with long videos, even when trained only with images}. Freezing the VLM and only tuning ReToken.}
\centering
\begin{subfigure}[t]{0.42\textwidth}
    \vspace{0pt}
    \caption{Performance gain for Qwen3VL on different splits of Video-MME.}
    \label{fig:videomme_gain}
    \hspace*{-2.7em}
    \includegraphics[height=.19\textheight]{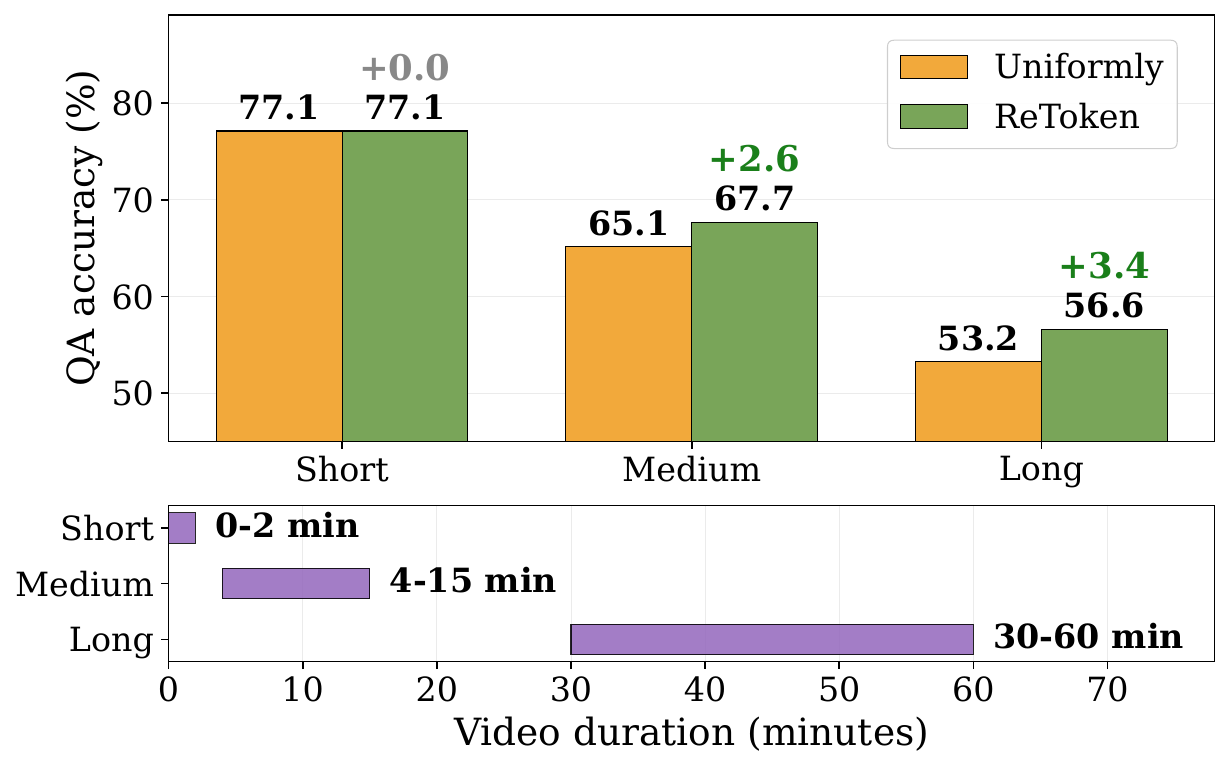}
\end{subfigure}
\hfill
\begin{subfigure}[t]{0.56\textwidth}
    \vspace{0pt}
    \caption{All results in the \texttt{Process raw frames} section are sourced from the technical report~\cite{bai2025qwen3}.}
    \label{tab:long_video_benchmarks}
    \centering
    \setlength\tabcolsep{1pt}
    \resizebox{\linewidth}{!}{%
    \begin{tabular}{@{}lc|cc}
    \toprule
    Model & \multirow{2}{*}{\# Frames}  & LVBench & Video-MME  \\
    Duration &  & 30 – 140 min  &  0 – 60 min  \\
    \midrule
    \multicolumn{4}{@{}l}{\textbf{\textcolor{gray}{\texttt{Process raw frames}}}} \\
    Gemini 2.5 Pro~\cite{comanici2025gemini}      & 512    & 69.0 & 80.6 \\
    OpenAI GPT-5~\cite{singh2025openai}           & 256    & -    & 77.3 \\
    Claude Opus 4.1~\cite{anthropic2025opus41card}& 100    & -    & 73.3 \\
    Qwen3VL-8B~\cite{bai2025qwen3}                & 2 FPS  & 58.0 & 71.4 \\
    \midrule
    \multicolumn{4}{@{}l}{\textbf{\textcolor{gray}{\texttt{Retrieve from the stored KV cache at 0.5 FPS}}}} \\
        
    Qwen3VL-8B~\cite{bai2025qwen3}                & 100 & 40.6 & 65.1 \\
    \quad \textbf{+ \textsc{ReToken}}                     & 100 
        & 48.6 {\color{softgreen}\text{\small(+8.0)}} 
        & 67.1 {\color{softgreen}\text{\small(+2.0)}} \\
    \bottomrule
    \end{tabular}}
\end{subfigure}
\end{table*}

%% file: text/5_ablations.tex
\subsection{Error Analysis and Ablations}
\label{sec:ablation}

\input{tab/ablation}


In this section, we analyze common failure modes of \modelname{} and ablate two design choices using \texttt{Qwen3VL-8B} on Visual Haystacks: (1) visual key- vs.\ value-based retrieval and (2) inference settings.


We further ablate several design choices in the supplement: token-only training versus partial tuning (Supp.~\ref{supp:supp_partial_tuning}), whether a single token suffices for retrieval (Supp.~\ref{supp:supp_multiple_tokens}), and whether \modelname can skip attending to visual tokens (Supp.~\ref{supp:supp_skip_attend_vision}). We also provide an efficiency analysis in Supp.~\ref{supp:supp_efficiency_analysis}.

\textbf{Error Analysis.} 
LVBench provides task-type labels, enabling us to analyze when \modelname{} helps or fails (Tab.~\ref{tab:lvbench_error_analysis}). Gains are largest when evidence is localized and nameable, and smaller when relevance depends on cross-frame or temporal structure rather than per-frame content. \modelname{} hurts on summarization, where evidence is dispersed across the video and uniform coverage is a better prior.

\begin{figure}[t!]
    \centering
    \makebox[\textwidth][c]{%
        \includegraphics[width=1\textwidth]{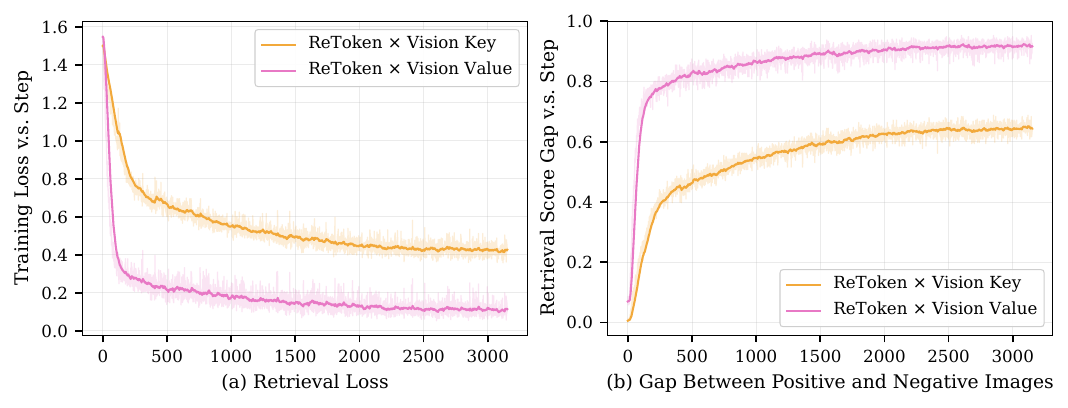}
    }
    \caption{\textbf{Training Comparison.} The \modelname$\times$Value variant (a)  converges faster in retrieval loss, and (b) learns a substantially larger retrieval score gap between relevant and irrelevant images.}
    \label{fig:compare_loss}
\end{figure}

\textbf{Retrieval Score Based on Average Vision Key or Value}. We ablate \modelname's scoring mechanism in Tab.~\ref{tab:retoken_key_value} with $C=50, K=1$: ``$\times$Key'' trains \modelname to retrieve via the average image keys, while ``$\times$Value'' uses the average image values. ``$\times$Value'' pulls clearly ahead in both recall and accuracy. We attribute this to values providing a more discriminative signal for distinguishing the relevant image among many distractors. Fig.~\ref{fig:compare_loss} provides further evidence for this. It shows the retrieval loss and retrieval score gap across training steps with the VLM frozen and only \modelname trained: $\times$Value yields a clear advantage.


\textbf{Single-Pass versus Two-Pass Inference}. 
We ablate the inference setting and report the accuracy in Tab.~\ref{tab:twp_pass_inference} with $C=50, K=1$. 
``Single'' means retrieving independently at each layer for generation, while ``Two'' lets early layers attend to all visual tokens and uses the final-layer retrieval for generation.
ReKV performs better with single-pass, likely because its attention-based retrieval directly supports answering, whereas \modelname is trained with retrieval supervision only at the final layer and benefits from two-pass inference. Thus, ReKV uses single-pass and \modelname{} two-pass inference by default.

%% file: tab/ablation.tex
\begin{table*}[b!]
\centering
\setlength{\tabcolsep}{4pt}

\begin{minipage}[t]{0.67\textwidth}
\vspace{0pt}
\centering

\caption{
\textbf{\modelname{} helps most when the evidence is localized and nameable,
and hurts when it is dispersed across the video.}
Per-type accuracy on LVBench. $K=100$.
\textit{\#QA} denotes the number of questions per type;
a question can carry multiple type labels.
}
\label{tab:lvbench_error_analysis}
\vspace{-0.4em}

\resizebox{\linewidth}{!}{%
\begin{tabular}{lcccr}
\toprule
Question type & \textit{\#QA} & Uniformly
& \textbf{\textsc{ReToken}} & $\Delta$ \\
\midrule
Key information retrieval & 291 & 42.3 & 57.7 & $+$15.4 \\
Entity recognition        & 677 & 40.0 & 50.5 & $+$10.5 \\
Reasoning                  & 201 & 40.8 & 46.3 & $+$5.5 \\
Temporal grounding         & 220 & 35.0 & 38.6 & $+$3.6 \\
Event understanding       & 647 & 41.4 & 43.3 & $+$1.9 \\
Summarization              & 58  & 36.2 & 31.0 & $-$5.2 \\
\bottomrule
\end{tabular}%
}

\end{minipage}
\hfill
\begin{minipage}[t]{0.29\textwidth}
\vspace{0pt}
\centering

\caption{Key vs.\ Value.}
\label{tab:retoken_key_value}
\vspace{-0.4em}

\setlength\tabcolsep{3pt}
\begin{tabular}{@{}l|cc@{}}
\toprule
& $\times$ Key & $\times$ Value \\
\midrule
Recall   & 59.4 & \textbf{64.7} \\
Accuracy & 70.6 & \textbf{72.0} \\
\bottomrule
\end{tabular}

\vspace{1.6em}

\caption{Two-pass Inference.}
\label{tab:twp_pass_inference}
\vspace{-0.4em}

\setlength\tabcolsep{4pt}
\begin{tabular}{@{}l|cc@{}}
\toprule
 & ReKV & \textbf{\textsc{ReToken}} \\
\midrule
Single & \textbf{51.2} & 50.4 \\
Two    & 50.4 & \textbf{72.0} \\
\bottomrule
\end{tabular}

\end{minipage}

\vspace{-0.5em}
\end{table*}

%% file: text/6_conclusion.tex
\section{Conclusion}

We diagnose the limitations of attention-based retrieval in VLMs and introduce \modelname, a learnable token that improves visual retrieval. Our diagnosis points to a broader principle: in pretrained VLMs, the value space carries a stronger text-aligned signal. Despite being trained only on multi-image data, \modelname yields significant improvements on both image and long-video benchmarks, transferring zero-shot from images to videos. Both training and long-video inference fit on a single H100, making \modelname a practical step toward scalable long-context multimodal reasoning.

\textbf{Limitations}. \modelname requires two forward passes and attends to more visual context in early layers, slightly increasing memory usage and response time. Our training data is limited to multi-image QA; training on video data may better capture temporal structure and improve video understanding.

\textbf{Future Work}. \modelname scores each frame independently using a frame-level representation, leaving several directions for improvement. (1) Some evidence is only clear across multiple consecutive frames, motivating retrieval of short frame sequences rather than individual frames. (2) Queries involving temporal relations, such as ``what happened before I entered the room,'' require locating the referenced event and retrieving frames before or after it. (3) Finally, token-level scoring could better capture small or localized visual evidence that may be lost through frame-level mean pooling.

\section*{Acknowledgments}

We thank Xiaodong Liu, Sethuraman T V, and Bolin Lai for their helpful discussions. This project has benefited from the Microsoft Agentic AI Research and Innovation (AARI) grant program, and was partially supported by the Office of Naval Research under grant N00014-23-1-2383. This work also used NVIDIA GPUs at NCSA Delta through allocation CIS240059 and CIS250059 from the Advanced Cyberinfrastructure Coordination Ecosystem: Services \& Support (ACCESS) program, which is supported by NSF Grants \#2138259, \#2138286, \#2138307, \#2137603, and \#2138296.

%% file: text/7_appendix.tex
\newpage
\appendix


\definecolor{allvisiontok}{RGB}{90, 175, 90}   
\definecolor{retvisiontok}{RGB}{240, 155, 50}    
\definecolor{questiontok}{HTML}{0071BC}
\definecolor{answertok}{RGB}{217, 130, 200}        

\newcommand{\allvis}{\mbox{\textcolor{allvisiontok}{All Vision Tokens}}}
\newcommand{\retvis}{\mbox{\textcolor{retvisiontok}{Retrieved Vision Tokens}}}
\newcommand{\rettok}{\mbox{\textcolor{retvisiontok}{<Retrieval>}}}
\newcommand{\qtok}{\mbox{\textcolor{questiontok}{Question Tokens}}}
\newcommand{\anstok}{\mbox{\textcolor{answertok}{Answer}}}

\newtcolorbox{promptbox}[1][]{
    enhanced,
    colback=gray!4,
    colframe=gray!55,
    colbacktitle=gray!20,
    coltitle=black,
    boxrule=0.5pt,
    arc=2pt,
    left=6pt, right=6pt, top=5pt, bottom=5pt,
    fonttitle=\bfseries\small,
    fontupper=\small\raggedright,
    title=#1
}

\section{Prompt Templates}
\label{supp:supp_prompts}

In this section, we detail the prompt templates used during training
and inference with \modelname. To highlight the
modifications introduced by \modelname, we color-code the
content-bearing tokens as follows:
\textcolor{allvisiontok}{All Vision Tokens},
\textcolor{retvisiontok}{Retrieved Vision Tokens},
\textcolor{questiontok}{Question Tokens} and the
\rettok{} token, and the model output \anstok{}.

\subsection{Original Prompt Template}

The standard prompt format used by vision-language models feeds
\emph{all} visual tokens into the model alongside the textual query:

\begin{promptbox}[Original Prompt]
<system><|vision\_start|>\allvis<|vision\_end|>\qtok<|im\_end|><|im\_start|>assistant: \anstok
\end{promptbox}

\subsection{Training Prompt Template}

During training, we employ two complementary prompt formats depending on the training mode. The first, which appends the \rettok{} token after the \textcolor{questiontok}{question} (teaching the model \emph{when} to invoke retrieval), is used in both frozen and partially fine-tuned settings:

\begin{promptbox}[Training Prompt 1: training the retrieval token]
<system><|vision\_start|>\allvis<|vision\_end|>\qtok \rettok
\end{promptbox}

The second format, designed to preserve the model's original QA abilities, is incorporated alongside the first exclusively when the LLM is partially fine-tuned:

\begin{promptbox}[Training Prompt 2: keeping its original QA ability]
<system><|vision\_start|>\allvis<|vision\_end|>\qtok<|im\_end|><|im\_start|>assistant: \anstok
\end{promptbox}

\subsection{Inference Pipeline}

At inference time, we adopt a two-pass \emph{retrieve-then-answer} pipeline. In Pass~1,  at most $K'$ frames are attended during retrieval and the model uses the \rettok{} token to aggregate query-relevant information, performing retrieval at the final layer. In Pass~2, only the retrieved visual tokens \retvis{} are supplied to the model to generate the final answer.

\begin{promptbox}[Inference Pass~1: retrieval triggering]
<system><|vision\_start|>\allvis<|vision\_end|>\qtok\rettok
\end{promptbox}

\begin{promptbox}[Inference Pass~2: answering with retrieved tokens]
<system><|vision\_start|>\retvis<|vision\_end|>\qtok<|im\_end|><|im\_start|> \\ assistant: \anstok
\end{promptbox}

\section{Additional Details}

\subsection{Training Data Filtering}
\label{supp:training_data}

The original MIRAGE fine-tuning set aggregates multiple open-source VQA datasets, many of which were already seen during Qwen3VL pretraining. We observe that on a substantial fraction of these examples, Qwen3VL achieves \emph{lower} generation loss
when conditioned on the full set of distractor images than on the target ground-truth image alone, suggesting that the model has memorized the underlying QA pairs and bypasses the intended retrieval task. To address this, we apply three filters to the MIRAGE-augmented dataset (in which images from different QAs are combined into a single multi-image input):

\begin{itemize}
\item \textbf{Memorization filter.} We discard examples where the
generation loss conditioned on the target image alone is higher than that
conditioned on the full image set, as such examples no longer require
retrieval to be answered.
\item \textbf{Difficulty filter.} We rank remaining examples by the
average attention score between the question query tokens and the
target image key tokens, and retain those with higher scores. In a
controlled experiment, we find that easier retrieval examples (higher
attention scores on the target image) lead to a stronger learned
retriever than harder examples.
\item \textbf{Multi-image filter.} We remove single-image QA examples (which lack distractor images for retrieval) and OCR-style queries (which test text recognition rather than visual recognition).
\end{itemize}

\input{tab/training_dataset}

After filtering, there are 70{,}686 examples in our final training dataset. We note that our filters are applied \emph{only} to the MIRAGE training split; the Visual Haystacks evaluation split and video datasets are left untouched. The filtering signals (generation loss and attention scores) are computed from the frozen Qwen3VL-8B on training examples only, with no access to evaluation labels or examples. Tab.~\ref{tab:training_data} shows the influence of filtering.

\subsection{Chain-of-Thought (CoT) Retrieval Baseline}
\label{supp:cot_prompts}

\begin{promptbox}[CoT Prompt Template (query-rewriting baseline)]
Given the following question, what phrase should I search for to find the visual evidence needed to answer it? Respond with the search phrase only, no more than 5 words.\\[2pt]
Question: \{question\}
\end{promptbox}

\subsection{How the Stored KV Cache Differs From Re-encoding the Visual Input}
\label{supp:KV_cache_distraction}

Higher retrieval recall does not ensure higher accuracy. We attribute this to
the causal-mask mechanism: later images always attend to earlier ones, so the stored KV cache of any given image becomes fused with information from preceding images. 

\input{tab/KV_cache_distraction}

To quantify this distraction, we compare the accuracy when using ``GT Image'' (re-encoding the ground-truth image alone, yielding a clean KV cache) with ``GT Cache'' (the fused KV cache for the ground-truth image). Their gap, $\Delta_{\text{GT Image} - \text{GT Cache}}$, measures the degree of distraction (Tab.~\ref{tab:KV_cache_distraction}); a smaller gap indicates a cleaner stored KV cache. Partial tuning on the retrieval dataset encourages the model to produce cleaner KV cache representations (KV cache is not affected when we only train the retrieval token). A cleaner KV cache benefits Visual Haystacks, but may pose a challenge for video understanding, which requires connecting information across adjacent frames.

\subsection{Runtime and Memory Usage}
\label{supp:supp_efficiency_analysis}

Tab.~\ref{tab:efficiency} breaks down per-question cost into retrieval and answer phases, and reports video encoding separately because encoding produces a persistent KV cache shared across all questions about the same video. Encoding dominates the budget at $\approx$14.7\,s per video. When few questions are asked of a video, the encoding dominates the compute time, but can be performed as a pre-process. \modelname adds roughly 0.4 seconds to the per-question retrieval and answering time, but significantly improves recall and question accuracy.

\input{tab/efficiency_analysis}

\section{Design Choices}
\label{supp:supp_design_choice}

\subsection{Token-Only Training versus Partial VLM Fine-Tuning}
\label{supp:supp_partial_tuning}

Tab.~\ref{tab:recall_acc_tradeoff} ablates the partial-tuning depth. Tuning the first few layers (1--3) gives the best accuracy while preserving high recall, indicating that early layers are where visual features can be best shaped without compromising downstream performance.

Tab.~\ref{tab:ablation_QAEgo4D} compares the transfer ability of the frozen and partial-tuning settings. \modelname shows a clear advantage in both. However, overall performance under partial tuning is lower than under the frozen setting. A possible reason is that tuning the LLM layers on image data can cause domain shift, especially given the distribution gap between image training data and video benchmarks.

\input{tab/tuning_ablations}

\subsection{Multiple Tokens}
\label{supp:supp_multiple_tokens}

Tab.~\ref{tab:multiple_tokens} reports results for training 3 ReTokens on the retrieval task. We append the 3 ReTokens sequentially, and compute the retrieval score by computing the cosine similarity between each projected ReToken and the averaged image value feature. We then average the resulting logits to obtain the final score.

Performance is on par overall, with small fluctuations in both directions. One reason may lie in our training recipe: the multi-token design treats the embeddings equally, since all three are supervised through one averaged score; nothing encourages them to specialize, so the tokens likely converge on a similar solution. A promising direction could be to empower specialization of different tokens so that a token is more pronounced for its specialized domain. But this is beyond the scope of this work and we leave it for future exploration.

\input{tab/multiple_tokens}

\subsection{Query Composition or Visual Summarization}
\label{supp:supp_skip_attend_vision}

Since we append the \modelname to the input, it can attend to both the images and the query, allowing it to serve as both a visual summarization token and a query summarization token. To better understand the role of the \modelname, we design an ablation in which the \modelname is only allowed to attend to the query tokens. In this setting, we perform two forward streams. In stream A, we encode the vision tokens and cache their KV. In stream B, we process only the query and the \modelname, so that neither the query tokens nor the \modelname can attend to the visual part. This way, the \modelname sees only the query tokens and learns to summarize what we want to retrieve based on the input query alone. We then compute the retrieval score by computing the cosine similarity between the projected \modelname from the last layer of stream B and the averaged visual value feature from the last layer of stream A. 

Tab.~\ref{tab:query_composition} reports the results with and without attention to the images. The results show that skipping the visual tokens entirely still yields decent performance, but underperforms the variant that attends to the images. This suggests that when the \modelname can understand what happens in the input video/images, it achieves better retrieval results.

\input{tab/query_composition}





%% file: tab/training_dataset.tex
\begin{table*}[t!]
\caption{\textbf{The influence of training dataset.} ``Unfiltered'' means directly sampling 70{,}686 examples from the original MIRAGE fine-tuning set without filtering, and using them to train ReToken. ``Filtered'' is our default setting.}
\label{tab:training_data}
\setlength\tabcolsep{7pt}
\centering
\resizebox{1\textwidth}{!}{
\begin{tabular}{@{}lcccccccc@{}}
\toprule
Method  & $C=1$ & $C=2$ & $C=3$ & $C=5$ & $C=10$ & $C=20$ & $C=50$ & $C=100$   \\

Qwen3VL-8B~\cite{bai2025qwen3}  & 87.8 & 82.0 & 77.9 & 74.7 & 69.2 & 64.0 & 58.6 & 55.7  \\

\quad Unfiltered & 87.8 & 85.2 & 83.1 & 80.8 & 78.1 & 75.5 & 69.3 & 62.7 \\

\rowcolor{SeaGreen!20}
\quad \textbf{Filtered} &  87.8 & 85.9 & 84.4 & 82.5 & 80.7 & 76.8 & 72.0 & 67.7 \\

\bottomrule 
\end{tabular}
}
\end{table*}

%% file: tab/KV_cache_distraction.tex
\newcommand{\g}{\cellcolor{SeaGreen!20}}

\begin{table*}[b!]
\caption{\textbf{Partial tuning on the retrieval dataset yields a cleaner KV cache.} $\Delta_{\text{GT Image} - \text{GT Cache}}$ measures the degree of distraction. A smaller gap indicates a cleaner stored KV cache.}
\label{tab:KV_cache_distraction}
\setlength\tabcolsep{5pt}
\centering
\resizebox{1 \textwidth}{!}{
    \begin{tabular}{@{}l | cc | c | cc | cc@{}}
    \toprule
    \multirow{2}{*}{Model }  &  \multicolumn{2}{c|}{Loss}  & \multirow{2}{*}{GT Image}   & \multicolumn{2}{c|}{GT Cache}  & \multicolumn{2}{c}{$\Delta_{\text{GT Image} - \text{GT Cache}}$}  \\  
     & Generation & Retrieval  &  & $C=2$ & $C=50$  & $C=2$ & $C=50$   \\

    \cmidrule{1-8}
    Qwen3VL-8B &  &  & 
    87.8 & 86.5 & 80.7 &  1.3 & 7.1 \\

    \noalign{\vskip 3pt}\cdashline{1-8}\noalign{\vskip 3pt}
    
    \multirow{2}{*}{Tuned Layer 1 - 3 }   & \checkmark & & 
    87.8 & 86.8 & 82.3 &  \textbf{1.0} & 5.5 \\


    & \checkmark & \checkmark  &  \textbf{88.5} & \textbf{87.3} &  \textbf{83.6}  &   1.2  &  \textbf{4.9} \\
    
    \bottomrule
    \end{tabular}
}
\end{table*}

%% file: tab/efficiency_analysis.tex
\begin{table}[H]
\setlength{\abovecaptionskip}{5pt}
\caption{\textbf{Runtime vs.\ quality on long-video QA} (QAEgo4D$_{\text{Test-MC}}$, $\sim$240 frames per video at 0.5 FPS, single H100). Latency is reported \emph{per question} after one-time video encoding. \textsc{ReToken} achieves higher accuracy with only modest overhead in the retrieval pass, and identical cost for the encoding and answer stages.}
\label{tab:efficiency}
\setlength{\tabcolsep}{5pt}
\centering
\resizebox{1\linewidth}{!}{%
\begin{tabular}{l c | c c c | c c | c c}
\toprule
&  & \multicolumn{3}{c|}{Latency (second, per question)} & \multicolumn{2}{c|}{Peak GPU (GB)} & \multicolumn{2}{c}{QA (\%)} \\
Method & $K$ & Encode$^{\dagger}$ & Retrieve / Load & Answer & Retrieve & Answer & Acc. & Recall \\
\midrule
Uniformly             &   16         & 14.68 &  0.081       & 0.169 & 65.1       & 65.4    & 53.6          & 38.0            \\
\textbf{\textsc{ReToken}}    &   16    & 14.68 & 0.519 & 0.167 & 66.6 & 65.4  & \textbf{60.0} & \textbf{70.6} \\
\midrule
Uniformly                       & 32 & 14.70 & 0.128       & 0.166 & 66.0       & 66.7    & 55.2          & 59.6            \\
\textbf{\textsc{ReToken}}       & 32 & 14.70 & 0.538 & 0.166 & 67.2 & 66.8  & \textbf{60.6} & \textbf{81.2} \\
\bottomrule

\end{tabular}
}
\\[2pt]
{\footnotesize $^{\dagger}$Encoding is performed once per video; answers are generated from the visual tokens of the $K$ retrieved frames.}
\end{table}

%% file: tab/tuning_ablations.tex
\begin{table}[H]
\caption{\textbf{Tuning early layers can improve performance on the image task, but it degrades performance on the video task}. (a) Tuning the first few VLM layers (Layer 1--3) gives the best recall--accuracy trade-off at $C{=}50, K=1$. (b) Training the VLM on images decreases accuracy on video; evaluated on QAEgo4D$_\text{Test-MC}$ with $K{=}16$.}
\label{tab:tuning_ablations}
\centering
\begin{subtable}[t]{0.5\textwidth}
    \caption{Train on images, evaluate on images.}
    \label{tab:recall_acc_tradeoff}
    \centering
    \setlength\tabcolsep{3pt}
    \resizebox{\textwidth}{!}{%
        \begin{tabular}{@{}l | ccccc@{}}
        \toprule
        Tuning Layers& Freezing & 1 & 1-3 & 1-10 & N \\
        \cmidrule{1-6}
        Recall & 64.7 & \textbf{68.3} & 68.2 & 58.2 & 64.3 \\
        Acc    & 72.0 & 73.3 & \textbf{75.0} & 72.6 & 71.3 \\
        \bottomrule
        \end{tabular}
    }
\end{subtable}
\hfill
\begin{subtable}[t]{0.48\textwidth}
    \caption{Train on images, evaluate on video.}
    \label{tab:ablation_QAEgo4D}
    \centering
    \setlength\tabcolsep{3pt}
    \resizebox{\textwidth}{!}{%
        \begin{tabular}{@{}l | ccc@{}}
        \toprule
        Strategy & Uniformly & ReKV~\cite{di2025streaming} & \textbf{ReToken} \\
        \cmidrule{1-4}
        Freezing & 53.6 & 57.8 & \textbf{60.0} \\
        Layer 1 - 3 & 50.0 & 55.2 & \textbf{58.0} \\
        \bottomrule
        \end{tabular}
    }
\end{subtable}
\end{table}

%% file: tab/multiple_tokens.tex
\begin{table*}[t!]
\setlength{\abovecaptionskip}{5pt}
\caption{\textbf{Multiple tokens perform on par with a single token.} Qwen3VL-8B.}
\label{tab:multiple_tokens}
\setlength\tabcolsep{7pt}
\centering
\resizebox{1\textwidth}{!}{
\begin{tabular}{@{}lcccccccc@{}}
\toprule
Method  & $C=1$ & $C=2$ & $C=3$ & $C=5$ & $C=10$ & $C=20$ & $C=50$ & $C=100$   \\

\midrule

3 tokens & 87.8 & 85.5 & 84.3 & 82.5 & 82.0 & 77.6 & 72.1 & 67.9 \\

\rowcolor{SeaGreen!20}
\textbf{1 token} &  87.8 & 85.9 & 84.4 & 82.5 & 80.7 & 76.8 & 72.0 & 67.7 \\

\bottomrule 
\end{tabular}
}
\end{table*}

%% file: tab/query_composition.tex
\begin{table}[H]
\setlength{\abovecaptionskip}{5pt}
\caption{\textbf{Allowing the ReToken to attend to images yields better results.} Question tokens also skip the visual tokens in the ``skip images'' setting. The VLM is frozen.}
\label{tab:query_composition}
\setlength\tabcolsep{7pt}
\centering
\resizebox{1\textwidth}{!}{
\begin{tabular}{@{}lcccccccc@{}}
\toprule
Method  & $C=1$ & $C=2$ & $C=3$ & $C=5$ & $C=10$ & $C=20$ & $C=50$ & $C=100$   \\

\midrule

Qwen3VL-8B~\cite{bai2025qwen3}  & 87.8 & 82.0 & 77.9 & 74.7 & 69.2 & 64.0 & 58.6 & 55.7  \\

\quad skip images & 87.8 & 83.4 & 78.9 & 76.2 & 73.7 & 68.5 & 63.1 & 62.0 \\

\rowcolor{SeaGreen!20}
\quad \textbf{attend images} &  87.8 & 85.9 & 84.4 & 82.5 & 80.7 & 76.8 & 72.0 & 67.7 \\

\bottomrule 
\end{tabular}
}
\end{table}